\pdfoutput=1

\documentclass[11pt]{article}

\usepackage[final]{coling}
\usepackage{multirow}
\usepackage{placeins}
\usepackage{comment}
\usepackage{censor}
\usepackage{times}
\usepackage{latexsym}
\usepackage{enumitem}
\usepackage{multicol} 
\usepackage{listings} 
\usepackage[T1]{fontenc}
\usepackage[utf8]{inputenc}
\usepackage{microtype}
\usepackage{inconsolata}
\usepackage{graphicx}
\usepackage{float}       
\usepackage{multirow}    
\usepackage{booktabs}    
\usepackage{verbatim}    
\usepackage{caption}     
\usepackage{subcaption}  
\usepackage{algorithmic}
\title{Zero-shot Slot Filling in the Age of LLMs for Dialogue Systems}

\author{ Mansi Rana \\ {\bf Kadri Hacioglu} \\ {\bf Sindhuja Gopalan} \\ {\bf  Maragathamani Boothalingam} \\
Uniphore \\
\texttt{\{mansi.rana,kadri.hacioglu,sindhuja,maragathamani\}@uniphore.com}
}
 
\begin{document}
\maketitle
\begin{abstract}
Zero-shot slot filling is a well-established subtask of Natural Language Understanding (NLU). However, most existing methods primarily focus on single-turn text data, overlooking the unique complexities of conversational dialogue. Conversational data is highly dynamic, often involving abrupt topic shifts, interruptions, and implicit references that make it difficult to directly apply zero-shot slot filling techniques, even with the remarkable capabilities of large language models (LLMs). This paper addresses these challenges by proposing strategies for automatic data annotation with slot induction and black-box knowledge distillation (KD) from a teacher LLM to a smaller model, outperforming vanilla LLMs on internal datasets by 26\% absolute increase in F1 score. Additionally, we introduce an efficient system architecture for call center product settings that surpasses off-the-shelf extractive models by 34\% relative F1 score, enabling near real-time inference on dialogue streams with higher accuracy, while preserving low latency.

\end{abstract}

\begin{table*}[t] 
\centering
\begin{minipage}{0.45\textwidth}  
  \centering
  \begin{tabular}{|l|c|}
    \hline
    \textbf{Training set} & \textbf{Samples} \\
    \hline
    \verb|Multi-domain data| & 13700 \\
    \verb|In-house telecom data| & 2179 \\
    \verb|In-house insurance data| & 9240 \\
    \hline
  \end{tabular}
  \caption{Training datasets for baseline fine-tuning.}
  \label{tab:baseline-train}
\end{minipage}%
\hspace{0.05\textwidth}  
\begin{minipage}{0.45\textwidth}  
  \centering
  \begin{tabular}{|l|c|}
    \hline
    \textbf{Test set} & \textbf{Samples} \\
    \hline
    \verb|Multi-domain| & 3432 \\
    \verb|Seen domain, Unseen source| & 550 \\
    \verb|Unseen domain: Finance| & 2522 \\
    \hline
  \end{tabular}
  \caption{Test datasets for baseline fine-tuning.}
  \label{tab:baseline-test}
\end{minipage}
\end{table*}

\section{Introduction}
Slot filling for product-centric business use cases involves extracting essential information from customer interactions such as inquiries, complaints or feedback, and organizing it into predefined slots like product name, issue type, customer details, and resolution status. This approach enables customer service teams to quickly identify the nature of problems, streamline responses, and enhance the overall customer experience by automating certain aspects of the process, resulting in faster and more efficient support. Slot filling also enables thorough real-time and after-call analysis by organizing and making key conversation details easily accessible. This approach is often used to enhance after-call summaries, thereby reducing the need for agents to spend time manually writing reports.

The motivation to implement Zero-Shot NLU \citep{Bapna2017Towards, Palatucci2009Zero}, for slot filling lies in addressing traditional limitations \citep{Mehri2021GenSF} like high time to value (TTV), reliance on labeled data, and costly iterative training. Zero-shot models \citep{Touvron2023LLaMA} allow for immediate deployment without prior training. However, applying these methods to conversational data is challenging due to ambiguity, lack of context, and interruptions. For instance, correctly mapping implicit mentions or resolving references like “she,” “he,” or “it” is difficult without strong contextual understanding. Conversational shifts, slot values spanning across multiple turns, and slang expressions further complicate this task. While LLMs have improved contextual understanding, they have also introduced new challenges in deployment, such as latency, concurrency, and maintaining cross-domain functionality \citep{Shi2023Adaptive}. 

Combining slot descriptions with a small set of example slot values improves the model’s ability to generalize across different domains, though its reliance on accessible and representative examples remains a limitation \citep {Shah2019Robust}. To deal with ambiguity, they were reformulated into concrete questions by \citet{Du2021QA}, but poorly framed questions can negatively impact accuracy in slot prediction. Prompting techniques \citep{Li2023Generative, Luo2023Zero} enhanced adaptability by offering explicit cues or feedback, and improved adaptability while increasing complexity and computational overhead. 

To reduce computational overhead, this task was treated as a joint problem by combining intent detection with slot filling \citep{Zhang2016Joint}, but errors in one task can adversely impact the other. Retrieval augmented generation (RAG) approaches demonstrate strong generalization in low data settings \citep{Glass2021Robust}, but the effectiveness of the retrieval mechanism can be a bottleneck. Contrastive learning techniques \citep{Wang2021Bridge, Zhang2023Hierarchical}, shared embedding spaces \citep{Siddique2021Linguistically, Shi2023Adaptive} enable adaptability across new domains, but have their own challenges such as demanding significant computational resources and good embedding quality.

Addressing these issues calls for innovative solutions that can bridge the gap between zero-shot adaptability and practical deployment in real-world applications. Our contribution in this is two-fold:
\begin{itemize}[itemsep=0pt, topsep=0pt, partopsep=0pt, parsep=0pt]
\item First, we propose a tailored data annotation strategy incorporating slot induction \citep{Nguyen2023Slot} followed by black-box knowledge distillation (KD) \citep{Wang2021zero, Nguyen2022Black, Finch2024Transforming}, that transfers knowledge from the teacher model without accessing its internal architecture or parameters. This fine-tuning approach results in a model that is both highly generalizable and robust to conversational data.
\item Secondly, we demonstrate that the performance of a standard off-the-shelf extractive model commonly used in products can be significantly improved by integrating it as a preprocessing step alongside a fine-tuned model like ours and applying slot-specific constraints. This layered approach achieves near real-time performance with enhanced accuracy, while maintaining low latency and outperforming standalone extractive or LLM methods.
\end{itemize}

\section{Approach}
In this section we describe our framework for zero-shot slot filling. We developed a data annotation strategy based on black-box KD and slot induction and an architecture that leverages an extractive model to improve the accuracy and latency of the fine-tuned model. Using black-box KD, knowledge from a large foundation model, in our case, a commercial LLM with over 70 billion parameters, is transferred to a relatively small model through fine tuning using predictions from larger model with significantly reduced requirements for human annotation. 
In this approach, we follow a slot induction approach (instead of using predefined slot names), where we instructed the teacher model to predict all possible slot label-value pairs from the input text. This approach enhances the model's ability to generalize well across domains. We also used context along with the input text in the instruction to make the model context-aware. The training dataset comprises context and input text of varying lengths, which enables configuration flexibility for inference. The input to the system can be a single turn or multi turn. The following steps are required for this process:
\begin{algorithmic}[1] 
    \STATE Creation of an annotated dataset using a large commercial LLM with more than 70 billion parameters
    \STATE Conversion of the annotated dataset into an instruction dataset
    \STATE Instruction fine-tuning of a smaller model
    \STATE Refining and aligning predictions with human annotations
\end{algorithmic}

\section{Data Collection}

\subsection{Annotation of data}
The raw dataset is a collection of call transcripts, capturing conversations between agents and customers, sourced from contact center interactions. A specific prompt for the LLM to facilitate the targeted ``slot filling'' task was developed (see Appendix~\ref{sec:annotation prompt}). For slot induction, the LLM is instructed to discover novel slot labels with each annotation request. Iteratively, we obtain a growing list of slot labels as new ones are discovered. To refine/align these annotations with human annotations, we also employ consistent annotation guidelines across the same datasets for human annotators.

Our dataset comprises three diverse and distinct sources:
1) a balanced, multi-domain dataset sourced externally, encompassing five external domains of banking, insurance, telecommunications, retail and healthcare retail in multiple English accents; 2) an in-house dataset from the telecommunications domain; and 3) an in-house dataset from the insurance domain. While the first dataset is externally sourced, the in-house datasets are collected and maintained internally by our organization. 
    
All data used in our study has been anonymized, i.e., real identifiers and sensitive information were systematically replaced with fictitious equivalents to preserve confidentiality and comply with privacy standards, while maintaining the complexity of these dialogues. Appendix section~\ref{sec:dataAnn} presents some examples of annotated data.

\subsection{Creation of instruction fine-tuning dataset}

After all the transcripts were annotated turn-by-turn by the LLM (teacher) and/or humans, with slot labels and their corresponding values, we created an instruction dataset that is used for fine-tuning a smaller student model. An instruction sample consists of two parts: 1. the instruction, comprising a brief description of the system, the task description, and the input as context or text to be used for slot filling, and 2. the response, comprising the slot labels and their corresponding values. See Appendix~\ref{sec:fine-tuning prompt} for the template used for creating instruction samples.

To allow the system to be robust to any particular strategy of slot-filling, for every turn in a transcript, we randomize the number of turns in the “context” and “text” to create an instruction sample. “Context” here refers to the text in the transcript preceding the text from which the slots are to be extracted. This way, the model is able to generalize to different lengths of input context and text. We also randomize the type and the number of “distractor slot labels”, which refer to slots that are not present in the input text but are used as distractors in the input query. This also serves as a data augmentation strategy for creating more training data, reflecting the same completion under different “context”, “text” and “distractor slot labels”. 
See Table \ref{tab:baseline-train} for an overview of training data where 
each sample is an instruction sample extracted from a turn in a transcript as explained above.

\subsection{Test Data}

To ensure our model's generalization across multiple domains without compromising training, we benchmark it using three distinct datasets. The first is a ``multi-domain'' dataset, which comprises transcripts from the same sources included in our training set. The second, labeled ``seen domain, unseen source'' belongs to the Telecommunications domain which is seen in training but originates from a different source. The third, labeled ``unseen domain'' belongs to the Finance domain, which is not represented in our training dataset. See Table~\ref{tab:baseline-test} for an overview of the test data, where each sample is an instruction sample extracted from a turn in a transcript.

After baseline fine-tuning, we further introduce two new internal datasets from different domains (healthcare and financial services). Table~\ref{tab:additional datasets} provides an overview of these datasets that are used in upcoming section~\ref{sec:Model Generalization} Model Generalization.

\begin{table}[H]
  \centering
  \begin{tabular}{|l|c|}
    \hline
    \textbf{Training set} & \textbf{Samples} \\
    \hline
    \verb|Healthcare domain| & 1846 \\
    \verb|Financial Services domain| & 2209 \\
    \hline
    \multicolumn{2}{c}{} \\  
    \hline
    \textbf{Test set} & \textbf{Samples} \\
    \hline
    \verb|Healthcare domain| & 8832 \\
    \verb|Financial Services domain| & 11487 \\
    \hline
  \end{tabular}
  \caption{Overview of additional training and test datasets for extended fine-tuning and testing.}
  \label{tab:additional datasets}
\end{table}

\section{Model Development and Evaluation}
\subsection{Fine-Tuning, Inference, Optimization}
Fine-tuning was performed on NVIDIA A10G GPUs, and the software ecosystem was primarily based on the Huggingface framework. To optimize the fine-tuning process, we employed several advanced techniques and libraries, including PEFT (parameter-efficient fine-tuning), QLORA (quantized low-rank adaptation), Accelerate, and DeepSpeed. For optimization, we used the AdamW optimizer. We used F1 as our primary metric for model selection. After fine-tuning, we implemented an efficient inference pipeline for evaluation using the open source vLLM (Virtual Large Language Model) library designed for efficient inference of LLMs \citep{kwon2023efficient}. Prior to our primary experiments, we performed a series of preliminary tests that focused on optimizing a select set of critical hyper-parameters. See Appendix~\ref{sec:setups} for a detailed overview of the hyperparameters and configurations used at the fine-tuning and inference stages.

\begin{table*}[h]
\centering
\small
\begin{tabular}{|l|ccc|ccc|ccc|c|}
\hline
\multirow{2}{*}{Model} & \multicolumn{3}{c|}{Multi-domain} & \multicolumn{3}{c|}{Seen domain, Unseen source} & \multicolumn{3}{c|}{Unseen domain} & \multicolumn{1}{c|}{Average} \\
\cline{2-11}
 & P & R & F1 & P & R & F1 & P & R & F1 & F1 \\
\hline
Mistral v0.3 & 0.45 & 0.44 & 0.44 & 0.82 & 0.79 & 0.80 & 0.41 & 0.26 & 0.32 & 0.52 \\
Llama 3 8B & 0.46 & 0.49 & 0.47 & 0.80 & 0.77 & 0.78 & 0.43 & 0.24 & 0.31 & 0.52  \\
Phi3-mini & 0.45 & 0.49 & 0.47 & 0.82 & 0.73 & 0.77 & 0.30 & 0.22 & 0.25 & 0.50 \\
Gemma 2B & 0.43 & 0.28 & 0.34 & 0.71 & 0.61 & 0.66 & 0.28 & 0.14 & 0.19 & 0.40 \\
\hline
\end{tabular}
\caption{Baseline performance (Precision, Recall, F1) for pretrained models over three datasets and their average}
\label{tab:model-performance}
\end{table*}

\subsection{Evaluation}
Given the generative nature of LLMs, the metrics based on ``exact match'' of ground truth to the model responses are inadequate. These metrics often penalize responses that are semantically correct but differ syntactically or lexically. To address this, we adopt a more flexible evaluation strategy using lenient matching, i.e., if the system response is partially correct or incomplete, it receives a credit. Despite this, in cases where the model responses are in normalized form (e.g., for dates or emails), we do not obtain a ``lenient match''. To address this, we applied inverse text normalization (ITN) to both the ground truths and responses before evaluation.

All metrics reported in this paper—lenient precision, recall, and F1 scores—are calculated after applying ITN. Henceforth, ``F1'' refers to this modified metric. We elaborate on the lenient metric with a few examples in Appendix~\ref{sec:lenientMetrics}.

\subsection{Baseline Model Performances}
In this section, we evaluate multiple foundational LLMs and report their performance without applying fine-tuning. Due to computational constraints, we prioritized model selection based on two key criteria: the model's ability to follow instructions for structured output, and its baseline performance. 

Results in Table~\ref{tab:model-performance} show that Mistral v03 and Llama 3 achieved identical average F1 scores, while Phi3-mini demonstrated competitive performance despite its smaller size. By contrast, the Gemma 2B model underperformed compared to the other LLMs. We observed that Mistral occasionally failed to generate valid JSON outputs, which occurred 2-3 times more frequently than Llama 3. Although Phi3 showed promising results, its primary focus on English and limited multilingual capabilities made it less suitable for our planned language expansions. Consequently, we selected Llama 3 8B model as the base of our subsequent experiments.
\subsection{Fine-Tuned Model Performances}
Our initial fine-tuning experiment utilized the training sets from three internal datasets to assess the improvements over baseline models. Table~\ref{tab:baseline-vs-fine-tuned} shows the substantial performance gains achieved through fine-tuning. These results indicate that smaller LLMs, when used out-of-the-box, are inadequate for slot filling tasks that demand extensive world knowledge and robust language understanding for generating structured outputs consistently. The significant performance boost underscores the critical importance of fine-tuning for specific domains. The mediocre performance of pretrained foundational models was notably enhanced by 25\% absolute to achieve acceptable results post fine-tuning. This relative difference between a mediocre performance of a ``generalist'' model and stellar performance of a ``domain/task expert'' model is achieved by fine-tuning.

\begin{table}[t]
\centering
\small 
\begin{tabular}{|l|l|c|}
\hline
\textbf{Model} & \textbf{Dataset} & \textbf{F1} \\
\hline
\multirow{5}{*}{Baseline} & Multi-domain & 0.47 \\
 &  Seen domain, Unseen source & 0.78 \\
 & Unseen domain & 0.31 \\
\cline{2-3}
 & Average & 0.52 \\
\hline
\multirow{5}{*}{Fine-tuned} & Multi-domain & 0.61 \\
 & Seen domain, Unseen source & 0.92 \\
 & Unseen domain & 0.78 \\
\cline{2-3}
 & Average & 0.77 \\
\hline
\end{tabular}
\caption{Comparison of F1 scores between baseline and fine-tuned versions of the Llama 3 8B LLM}
\label{tab:baseline-vs-fine-tuned}
\end{table}

\subsection{Model Alignment}
Human annotated data can often serve as high-quality reference for fine-tuning NLP models, allowing models to learn from and align with human behavior. Additionally, human annotated data can be used to identify and reduce potential inconsistencies and noise in the training data and models' outputs, to improve accuracy and consistency. To evaluate the impact of human annotations, we perform fine-tuning with only human annotations, and then also with both LLM-generated and human-labeled data. Table~\ref{tab:ft-strategies} presents the results, showing the average F1 score improvement across the three test datasets previously considered.
\begin{table}[h]
\centering
\begin{tabular}{|l|c|}
\hline
\textbf{Fine-tuning Strategy} & \textbf{Avg F1} \\
\hline
Human Annotations Only & 0.74 \\
LLM Annotations Only  & 0.77 \\
Both Annotations & 0.78 \\
\hline
\end{tabular}
\caption{Comparison of fine-tuning results with and without human annotations}
\label{tab:ft-strategies}
\end{table}

\begin{table}[t]
\centering
\begin{tabular}{|l|c|}
\hline
\textbf{Experiment} & \textbf{Avg. F1} \\
\hline
Baseline & 0.73 \\
Fine-tuning (+ new datasets) & 0.74 \\
\hline
\end{tabular}
\caption{Generalization Experiment Result}
\label{tab:generalization-results}
\end{table}

The baseline model trained solely on human annotations significantly under-performed compared to the model trained exclusively on LLM annotations. This is in line with the empirical analysis we conducted on human annotations compared to LLM-annotations over a subset of 20 transcripts.

Humans are good with named entities and often create more specific labels, like breaking down ``monthly insurance cost'' into ``Old Monthly Cost'' and ``New Monthly Cost''. However, they miss out on abstractive slots like reason for call, product mentions, survey participation, etc. We quantified the number of missed labels—those not identified or annotated in the transcript—and found that human annotators tend to miss twice as many labels compared to LLMs. Specifically, human annotators missed an average of 8 labels in a transcript, whereas LLMs only missed 4. This performance gap highlights the broader knowledge transfer achieved by LLMs compared to human annotations.

Combining both annotations resulted in a modest 1\% absolute improvement, which we term as ``human alignment''. The results suggest that the benefit of fine-tuning on human plus machine-generated data is due the volume of added data, not higher quality of the labels. These findings imply the costly and time-consuming human annotation and alignment processes could potentially be bypassed, with only a modest sacrifice in accuracy.

\subsection{Model Generalization} \label{sec:Model Generalization}
We evaluated the model's performance on additional datasets summarized in Table~\ref{tab:additional datasets} without further fine-tuning, establishing a baseline. The difference between this baseline and the performance after fine-tuning with these datasets demonstrated the ``generalization gap''. 

Table~\ref{tab:generalization-results} summarizes the result using F1 scores averaged across our five test sets. The final result demonstrates the model's generalization capability. The increase in average F1 score from 0.73 to 0.74 indicates a marginal generalization gap, demonstrating strong generalization capability of the model on other domains. 

\begin{figure}
    \centering
    \includegraphics[width=1\linewidth, height=7cm]{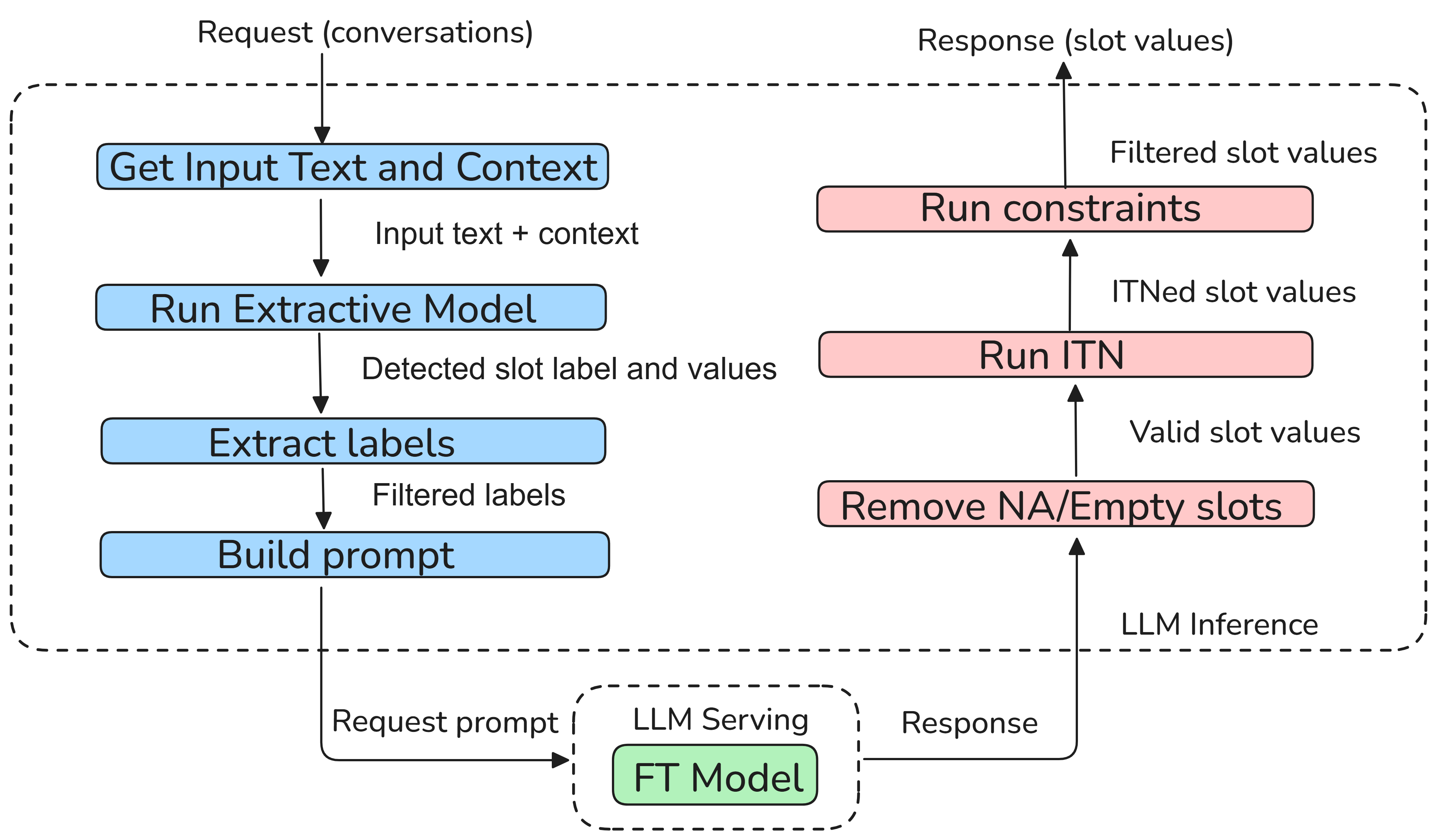}
    \caption{System Architecture: The LLM Inference module handles both preprocessing and postprocessing tasks. The fine-tuned (FT) LLM model is served by LLM Serving module. }
    \label{fig:System Architecture}
\end{figure}

\section{System Architecture}
Figure~\ref{fig:System Architecture} describes our system architecture. The two main components in our architecture are LL-inference and LL-serving. LLM-inference handles the pre/post processing of the data, while LLM-serving serves the (LLM) model. Once the system receives a conversation during call processing, the preprocessing module selects the input text and context. This text is passed to an extractive model, specifically GLiNER \citep{Zaratiana2024GLiNER}, which extracts slot values. GLiNER excels in recall for extracting slot values but lacks precision and cannot extract abstractive slots (e.g., Call Reason or Claim Issue). However, GLiNER can identify whether the conversation pertains to these labels. Slot values extracted by the lightweight model (GLiNER) are used to narrow down the list of slots requested from the LLM.

A prompt is constructed containing the input text, context, and the pre-determined, reduced set of slot labels by GLiNER. This prompt is sent to the fine-tuned LLM, and the LLM's response is processed to eliminate any empty or NA responses. Subsequently, ITN is applied to the results to filter out false positives and improve precision using constraints. A constraint is a predefined rule or user-defined parameter that is designed to minimize erroneous value extractions by zero-shot models. Predefined constraints are set for each entity type, while user-defined constraints apply to individual slots. Predefined constraints include matching entity types like date, duration, cardinal, money, email and standard named entities, whereas user-defined constraints, like lengths of values, can be applied to the numerical and alphanumeric entities (e.g., Dosage slot should have “partial cardinal” as a constraint).

\begin{table}[t]
\centering
\begin{tabular}{|l|c|}
\hline
\textbf{Method} & \textbf{Gain \%} \\
\hline
Only GLiNER & 2\% \\  
GLiNER + Constraints & 16\% \\
GLiNER + ft-llm + Constraints & 34\% \\
\hline
\end{tabular}
\caption{Comparison of different pipelines of our system over the legacy product baseline on unseen data.}
\label{tab:product-performance}
\end{table}

Table~\ref{tab:product-performance} compares the percentage gains in F1 scores compared to current product's baseline, which is a legacy extractive model. This table shows the incremental improvements achieved at each stage of system enhancement. Initially, off-the-shelf GLiNER model delivers a modest 2 \% increase in the F1 score. However, with the introduction of a constraints stage, this improvement becomes more pronounced, boosting the score by 16 \%. Notably, integrating GLiNER as a preprocessing step before the fine-tuned large language model (ft-LLM) and applying constraints results in a substantial 34\% improvement in the F1 score.

More details on the system performance are provided in section~\ref{sec:systemPer}. This staged approach demonstrates how integrating the extractive model as a preprocessing module before using the fine-tuned LLM, and applying constraints as postprocessing can lead to substantial improvements in both accuracy and reliability for complex industrial applications.  
 
\section{Limitations and Future Work}

\subsection{Expansion to other languages}
There are ongoing experiments to expand our approach to multiple languages including Spanish, Hindi, French, German and Arabic. Spanish and Hindi experiments indicate that the performances of vanilla Llama3 8b model trained with each language separately are comparable to those of a single fine-tuned model trained with all languages. Thus, we illustrate the possibility of a single multi-lingual model capable of slot filling in multiple languages. 

\subsection{Exploration of smaller models}
Initial experiments involving much smaller models like Phi3-mini have demonstrated that there is room for improvement by training smaller and compute-efficient models that exhibit enhanced capabilities. These advancements not only hold the promise to support a wider array of languages but also enable faster inference and reduced computational cost. 

\subsection{More robust evaluation metrics }
We discussed how the current evaluation metrics do not fully capture semantics and are not correlated with our small-scale human evaluations, underestimating model performance. Future work could benefit from adopting more form and content aware evaluation metrics. Ongoing work considers 1) weighted average of lenient F1 scores, ROUGE and BERTScores \citep{zhang2020bertscore}, and 2) slot-type specific evaluation using relevant metrics, such as ROUGE for form-insensitive slots and BERTScore for semantic slots.
  
\section{Conclusion}
In this paper, we have proposed a comprehensive, practical, and scalable approach for high-performance zero-shot slot filling using black-box knowledge distillation for conversational data. Through comprehensive experimentation, we demonstrated the effectiveness of using a larger LLM (teacher model) for creating data and transferring knowledge (through fine-tuning) to smaller LLMs (student models). In addition, we showed that fine-tuning significantly improved domain-specific performance, with the Llama 3 8B model outperforming the other foundation models we explored, achieving a 26\% absolute improvement in F1 over its vanilla version.
We also demonstrated a flexible and scalable deployment architecture supporting multiple use cases, including agent assistance, automated self-service, and post-call analytics. 
We used preprocessing with off-the-shelf GliNER model and postprocessing with slot constraints to improve the baseline system performance by 34\% relative F1. We highlighted future work for language expansion, the use of more efficient smaller models and  developing human-correlated metrics to better assess real-world performance. This work contributes to the growing body of research on practical applications of LLMs in customer service domains and provides a practical foundation for future developments in this field.
\newpage
\section{Acknowledgements}
We sincerely thank Andreas Stolcke, Roberto Pieraccini, Aravind Ganapathiraju, Malolan Chetlur and Neha Gupta for their valuable discussions that greatly influenced this work. We thank the product team at Uniphore for their product related insights. We also express our appreciation to Manickavela Aruguman for throughput and latency experiments.
\bibliography{custom.bib}
\section{Appendix}
\subsection{Annotation Prompt}
\label{sec:annotation prompt}
Following is the prompt template with variables “labels” and “text”, which are filled with a seed list of slot names (that reflects our knowledge/experience of slot labels across several domains) and the call transcripts, respectively:
\begin{quote}
\begin{lstlisting}[breaklines=true, basicstyle=\small]
You are an expert in Natural Language Processing.

Your task is to identify all named slot values in the given dialogue text, in which agent turn starts with ``Agent says:'' and customer turn starts with ``Customer says:''. 

Return the output in a json format for every line in the dialogue where key is text and value is dict of slot types and values. If there are no slot types in the line, return NA.

To get started, here is the list of slot types available to you: {labels}. 

Do not be restricted by this list. You should also extract slot types that are not in this list but present in the text.

Dialogue Text: {text}

\end{lstlisting}
\end{quote}

\subsection{Data Annotation Examples} \label{sec:dataAnn}
Following are some examples of labeled utterances derived from our dataset. These examples illustrate typical interactions in a call center setting, along with the corresponding labeled information (slots) extracted from the dialogue.
\begin{verbatim}
Example 1: 
Text: "Thank you for calling Net Company. 
        How can I assist you today?"
Slots: {"Company Name": "Net Company"}
\end{verbatim}
\begin{verbatim}
Example 2: 
Text: "Yes, uh I'm John Doe, and the 
        account number is 123456. 
        My wifi doesn't work."
Slots: {"Customer Name": "John Doe", 
        "Account Number": "123456", 
        "Reason for call": "wifi
         doesn't work"}
\end{verbatim}

\subsection{Fine-tuning Prompt}
\label{sec:fine-tuning prompt}
Following is the template for creating training samples for the instruction fine-tuning task:
\begin{quote}
\raggedright
\begin{lstlisting}[breaklines=true, basicstyle=\small]
<s>[INST]<<SYS>>
You are an honest and helpful information extractor.
<</SYS>> 
Your task is to extract values for the following slot labels in the Main Text delimited by triple backticks: {target slot labels}, {distractor slot labels}. Format your response as a JSON object with slot labels as the keys and slot values in a list. Only return the slots found the Main text. Use the following dialogue only as context support to extract slots from the Main text delimited by triple backticks:
{context text}
```
Main text: {text}
```
[/INST] {completion text}
\end{lstlisting}
\end{quote}
The variables in this template are: 
\begin{flushleft}
\resizebox{0.45\textwidth}{!}{%
\begin{tabular}{@{}p{.45\columnwidth}p{.55\columnwidth}}
{\em domain} & domain of the call transcript\\
{\em target slot labels} & labels in the text of interest\\
{\em distractor slot labels} & labels that don't exist in text\\
{\em context text} & the textual information that precedes the text\\
{\em text} & the textual information to be processed for slot filling of target slot labels\\
{\em completion text} & the response part of the prompt that the model will be trained to generate given all the information between the tags [INST] and [/INST] \\
\end{tabular}%
}
\end{flushleft}

\subsection{Fine-tuning and Inference Setups}
\label{sec:setups}
The table presents a summary of the hyper-parameters and configuration settings used in our fine-tuning steps, including hardware specifications, LORA settings, optimization parameters, and training details.
\begin{table}[h]
\centering
\setlength{\tabcolsep}{3pt} 
\begin{tabular}{|l|r|}
\hline
\textbf{Parameter} & \textbf{Value} \\
\hline
GPUs & 4 \\
GPU Memory & 24GB per GPU \\
LORA Rank & 16 \\
LORA Alpha & 32 \\
Dropout Rate & 0.05 \\
Batch Size per GPU & 1 \\
Gradient Accumulation Steps & 4 \\
Effective Batch Size & 16 \\
Maximum Learning Rate & 2e-4 \\
Number of Epochs & 5 \\
Warm-up & 10\% of iterations \\
AdamW beta1 & 0.9 \\
AdamW beta2 & 0.999 \\
AdamW epsilon & 1e-8 \\
Weight Decay & Not applied \\
Gradient Clipping Threshold & 1.0 \\
Adaptation Layers &  All linear layers \\
\hline
\end{tabular}
\caption{Fine-tuning Hyperparameters and Configuration}
\end{table}
Prior to our main experiments, we performed a series of preliminary experiments for optimizing a select set of critical hyperparameters. Specifically, we examined the effects of varying the number of training epochs, the size of the training dataset, the rank of the LORA (Low-Rank Adaptation) matrices, and the neural network layers subjected to fine-tuning. 
After the fine-tuning process, we implemented an efficient inference pipeline to evaluate our fine-tuned models using compute efficient vLLM inference engine. We have used greedy search with temperature 0. The maximum number of new tokens was set to 512. This temperature setting without sampling is particularly useful when we want consistent, high-confidence responses from the model. The choice of 512 tokens allows for reasonably lengthy responses while preventing excessively long generations.

\subsection{Lenient Metric Examples}
\label{sec:lenientMetrics}
In this section, we present some examples of lenient metrics for slot matching in dialogue systems. While traditional extractive systems select spans from user utterances, modern generative systems may rephrase or reformulate the extracted information, making lenient matching particularly crucial. The following examples demonstrate how generative systems could produce variations of the same semantic content, requiring more flexible evaluation metrics compared to exact span matching used in extractive approaches. Lenient matching allows for partial matches and semantic equivalence, providing a more realistic evaluation of slot filling systems compared to strict matching. Consider the following reference slot values:

\begin{verbatim}
Reference: {
  "time": "7:00 PM",
  "people": "2 people",
  "restaurant": "Joe's Pizza & Italian 
                Restaurant"
}
\end{verbatim}
In the following, we analyze two different predictions under both strict and lenient matching criteria:

\begin{verbatim}
Prediction 1: {
  "time": "7 PM",
  "people": "two",
  "restaurant": "joes pizza"
}
Prediction 2: {
  "time": "19:00",
  "people": "couple",
  "restaurant": "Joe's Italian Restaurant"
}
\end{verbatim}

Under strict matching criteria:
\begin{itemize}
    \item Prediction 1: 0/3 correct (no exact matches)
    \item Prediction 2: 0/3 correct (no exact matches)
\end{itemize}

Under lenient matching criteria:
\begin{itemize}
    \item Prediction 1: 3/3 correct
    \begin{itemize}
        \item ``7 PM'' matches ``7:00 PM'' (time format variation)
        \item ``two'' matches ``2 people'' (numerical equivalence)
        \item ``joes pizza'' matches ``Joe's Pizza \& Italian Restaurant'' (partial name match, missing punctuation)
    \end{itemize}
    \item Prediction 2: 3/3 correct
    \begin{itemize}
        \item ``19:00'' matches ``7:00 PM'' (time format equivalence)
        \item ``couple'' matches ``2 people'' (semantic equivalence)
        \item ``Joe's Italian Restaurant'' partially matches ``Joe's Pizza \& Italian Restaurant'' (partial name match)
    \end{itemize}
\end{itemize}

The lenient matching implementation involves:

\begin{enumerate}
    \item \textbf{Time normalization:} Converting different time formats to a standard representation
    \item \textbf{Numerical equivalence:} Matching different representations of numbers (words, digits)
    \item \textbf{Name normalization:} 
    \begin{itemize}
        \item Handling missing punctuation (Joe's vs Joes)
        \item Handling partial matches (subset of full name)
        \item Handling special characters (\& vs and)
        \item Case-insensitive matching
    \end{itemize}
    \item \textbf{Semantic match:} Using word embeddings or knowledge bases for semantic equivalence
\end{enumerate}

It should be noted that in all our metrics presented in this paper  we have not employed semantic similarity. However, in our limited internal experimentation, we have seen that the scores are further elevated with semantic similarity reflecting anecdotal human judgments better. 

\subsection{GLiNER}
GLiNER is a compact NER model designed to efficiently extract various types of entities from text. Unlike larger autoregressive models, GLiNER uses a bidirectional language model to process text and extract entities. It uses bidirectional transformer encoder to perform parallel entity extraction, making it more efficient than sequential models. The general approach is to place both span and entity embedding in the same latent space and then assess their compatibility, enabling accurate identification of entities. It outperforms in zero-shot NER scenarios on multiple NER benchmarks. It is more efficient, scalable and versatile approach to NER.

\subsection{System Framework}
In our architecture, we have designed two key services to efficiently handle the zero-shot slot-filling system using an LLM for call center tasks. These services operate as: 
\begin{enumerate}[itemsep=0pt, topsep=0pt, partopsep=0pt, parsep=0pt]
\item \textbf{LLM-Inference Service}: This service is responsible for data preprocessing and postprocessing. It ensures that input data is properly formatted and contextualized before being passed to the LLM. Additionally, it manages GLiNER model integration for extractive tasks. GLiNER is utilized here to extract relevant slot values from the conversation, narrowing down the slots for the LLM to process. 
\item \textbf{LLM-Serving Service}: This service focuses on serving the LLM model itself. It directly handles the inference requests and provides outputs based on the preprocessed data from the LLM-Inference service. 
\end{enumerate}

\subsection{System Deployment}
Both services (LLM-inference and LLM-serving) are deployed as Docker containers on \textbf{AWS}, taking full advantage of cloud-based infrastructure for scalability, reliability, and flexibility. The model serving is built using a combination of \textbf{Transformers} and \textbf{Seldon Core} libraries, ensuring robust performance and flexibility for various LLM use cases.

\subsection{System Performance} \label{sec:systemPer}
To optimize the system for both \textbf{latency} and \textbf{throughput}, we benchmarked it across different configurations, including FP16 precision, GPTQ-4bit, and GPTQ-8bit. Among these, the \textbf{FP16 model} with \textbf{prefix caching} demonstrated the best trade-off between performance and resource utilization. The use of prefix caching significantly improved both concurrency and throughput. By caching common portions of input sequences, redundant computations during inference can be reduced, leading to a notable reduction in processing times. With the \textbf{FP16 + prefix caching} setup, we achieved a concurrency level of \textbf{100 requests} while maintaining an average latency of \textbf{3.8 seconds} on an \textbf{NVIDIA L40 GPU}. This level of performance ensures that the system can handle near real-time, high-volume call center conversations with minimal delay, improving the overall user experience.

\FloatBarrier

\end{document}